\documentclass{article}
\usepackage[T1]{fontenc}

\usepackage[english]{babel}
\addto\captionsenglish{%
}

\usepackage[letterpaper,top=2cm,bottom=2cm,left=3cm,right=3cm,marginparwidth=1.75cm]{geometry}

\usepackage[numbers,sort&compress,square]{natbib}
\usepackage{amsmath}
\usepackage{graphicx}
\usepackage{booktabs}
\usepackage{multirow}
\usepackage{xcolor}
\usepackage[colorlinks=true, allcolors=blue]{hyperref}
\usepackage{parskip}
\usepackage{enumitem}
\usepackage{tikz}
\usetikzlibrary{positioning}
\usepackage{amssymb}  
\usepackage{pifont}   
\usepackage{array}
\usepackage{longtable}  
\usepackage{listings}
\usepackage{caption}
\usepackage{tcolorbox}

\date{}
\title{Holistic Evaluation and Failure Diagnosis of AI Agents}
\author{
  {\small Netta Madvil, Gilad Dym, Alon Mecilati, Edo Dekel,}\\
  {\small Jonatan Liberman, Rotem Brazilay, Liron Schliesser, Max Svidlo,}\\
  {\small Shai Nir, Orel Shalom, Yaron Friedman, David Connack,}\\
  {\small Amos Rimon, Philip Tannor, and Shir Chorev}\\[6pt]
  {\small Deepchecks, Ramat Gan, Israel}
}
\begin{document}
\maketitle
\begin{abstract}
AI agents execute complex multi-step processes, but current evaluation falls short: outcome metrics report success or failure without explaining why, and process-level approaches struggle to connect failure types to their precise locations within long, structured traces. 
We present a holistic agent evaluation framework that pairs top-down agent-level diagnosis with bottom-up span-level evaluation, decomposing analysis into independent per-span assessments. This decomposition scales to traces of arbitrary length and produces span-level rationales for each verdict. 
On the TRAIL benchmark, our framework achieves state-of-the-art results across all metrics on both GAIA and SWE-Bench, with relative gains over the strongest prior baselines of up to 38\% on category F1, up to 3.5$\times$ on localization accuracy, and up to 12.5$\times$ on joint localization-categorization accuracy. Per-category analysis shows our framework leading in more error categories than any other evaluator. 
Notably, the same frontier model achieves several times higher localization accuracy when used inside our framework than as a monolithic judge over the full trace, showing that evaluation methodology, not model capability, is the bottleneck.
\end{abstract}

\section{Introduction}

AI agents are increasingly deployed in production, powering customer assistants, developer tools, and automated workflows that take real actions on behalf of users. As autonomy grows, so do the consequences of failure: incorrect outputs, wasted tool budgets, harmful side-effects, and eroded user trust. Reliable evaluation is therefore essential, both for catching failures before deployment and for diagnosing where and why an agent went wrong so it can be improved. Agent executions are captured by tracing infrastructure, e.g. OpenTelemetry~\citep{opentelemetry2024genai}, as a hierarchical \emph{trace} of \emph{spans}, each a unit of activity such as an LLM call or tool invocation.

Most existing evaluation approaches focus on end-to-end outcomes~\citep{zheng2023judging, zhuge2024agent}. Benchmarks such as GAIA~\citep{mialon2024gaia} and AgentBench~\citep{liu2024agentbench} score agents primarily on task success or final-answer correctness, telling developers \emph{whether} an agent succeeded but not \emph{where} or \emph{why} it failed. More granular approaches have emerged that target intermediate behavior, most directly TRAIL's LLM-judge protocol~\citep{deshpande2025trail}, Agent GPA~\citep{datta2025gpa}, and AgentCompass~\citep{kartik2025agentcompass}, all of which target error identification on the TRAIL benchmark. However, reliably connecting failure types to their precise locations within long, structured traces remains difficult: monolithic LLM-judges scale poorly to long executions, while multi-stage pipelines localize better but struggle to name failure types reliably.

\begin{figure}[ht]
    \centering
    \includegraphics[width=\textwidth]{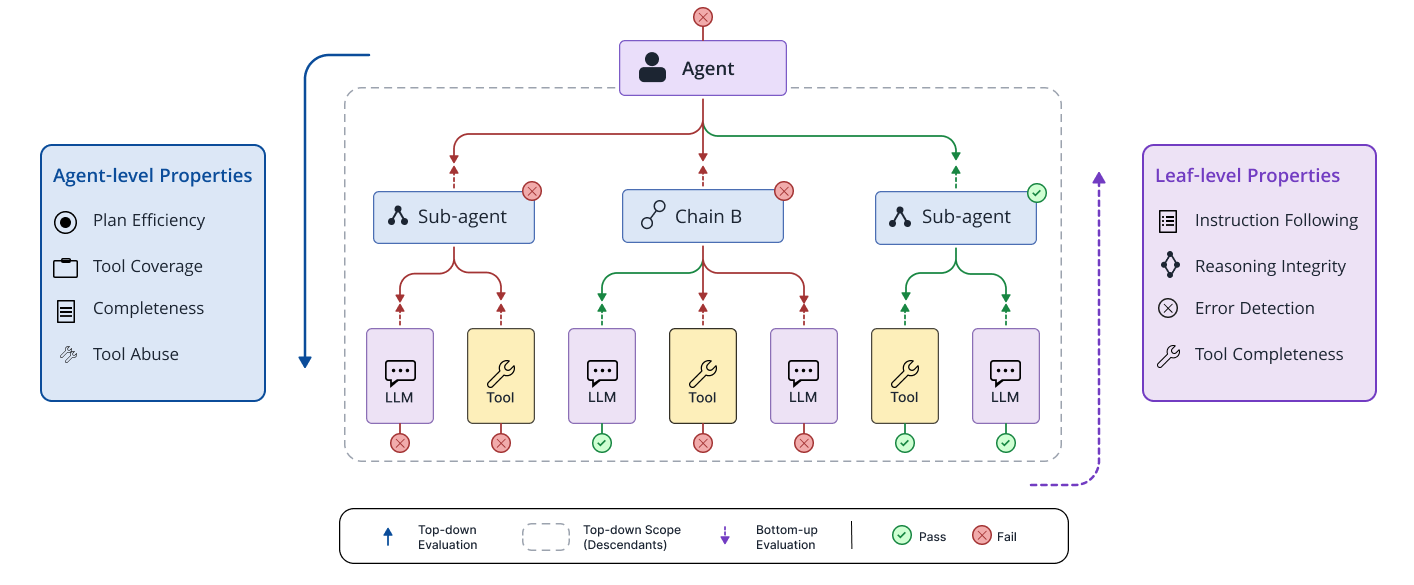}
    \caption{\textbf{\small  Overview of the holistic agent evaluation framework.} \small Top-down evaluation assesses agent-level metrics by reasoning over an agent span's descendants, while bottom-up evaluation localizes and categorizes failures at individual LLM and tool spans.  }
    \label{fig:framework}
\end{figure}

Our framework integrates complementary top-down and bottom-up perspectives, illustrated in Figure~\ref{fig:framework}. Top-down evaluation assesses agent-level metrics such as planning quality and tool coverage by analyzing an agent's descendants; bottom-up evaluation inspects individual spans (e.g. LLM calls, tool invocations) to localize failures and categorize their causes. Neither view alone suffices: bottom-up signals lack global context, while top-down signals lack local precision. By decomposing evaluation into independent span-level assessments, our framework scales to long traces and yields focused, accurate per-span verdicts. Trace-level assessments complement these by capturing behavioral patterns such as redundant calls, plan deviation, and incomplete coverage, which emerge only across spans.

We evaluate our framework on the TRAIL benchmark~\citep{deshpande2025trail} across both its GAIA~\citep{mialon2024gaia} and SWE-bench~\citep{jimenez2024swebench} datasets, leading on localization accuracy, weighted category F1, and joint localization-categorization accuracy. As traces grow longer, every prior baseline's localization accuracy degrades sharply or fails outright while our framework's holds steady. Notably, GPT-5.4 achieves 2.8$\times$ higher localization accuracy on GAIA and 12$\times$ higher on SWE-bench inside our framework than as a monolithic judge, demonstrating that the bottleneck is methodology, not model capability.

\paragraph{Contributions.}\mbox{}\\
Our main contributions are:
\begin{enumerate}
    \item A holistic, trace-aware agent evaluation framework that combines top-down agent-level assessment with bottom-up span-level analysis for fine-grained failure diagnosis.
    \item A span-level failure diagnosis method that localizes errors within agent traces, categorizes their causes, and provides natural language rationales.
    \item A broad analysis of the TRAIL benchmark, where our framework achieves the best results across all three metrics on both GAIA and SWE-Bench, with relative gains over the strongest prior baselines of up to 38\% in category F1, 3.5$\times$ in localization accuracy, and 12.5$\times$ in joint localization-categorization accuracy. The gains are consistent across datasets: 26\% / 1.25$\times$ / 2.58$\times$ on GAIA and 38\% / 3.5$\times$ / 12.5$\times$ on SWE-bench, respectively.
\end{enumerate}

\section{Related Work}

\paragraph{Agent evaluation datasets.} Outcome-only benchmarks 
(GAIA~\citep{mialon2024gaia}, 
SWE-bench ~\citep{jimenez2024swebench}, 
AgentBench~\citep{liu2024agentbench}, 
WebArena~\citep{zhou2024webarena}, 
AgentBoard~\citep{ma2024agentboard}) evaluate task success via 
final-answer correctness or progress against subgoals, without 
fine-grained error annotations. Process-annotated datasets vary 
in scope: Who\&When~\citep{zhang2025whoandwhen} annotates 127 
multi-agent failure logs with responsible agent and decisive 
step but no error-type taxonomy; 
AgentRx~\citep{barke2026agentrx} annotates 115 failed 
trajectories from $\tau$-bench, Flash, and Magentic-One under a 
9-category taxonomy but excludes successful traces; 
AgentRewardBench~\citep{lu2025agentrewardbench} annotates 1{,}302 
web-agent trajectories for LLM-judge alignment without 
fine-grained error categories; MAST~\citep{cemri2025mast} 
contributes a 14-mode taxonomy from 200 multi-agent traces.
We adopt TRAIL~\citep{deshpande2025trail}, which annotates 148
OpenTelemetry~\citep{opentelemetry2024genai} traces from GAIA
(multi-agent Open Deep Research) and SWE-bench (single-agent CodeAct)
with 841 span-level errors under a 20+ category taxonomy. To the best
of our knowledge, TRAIL is the only publicly available
process-annotated benchmark that combines span-level granularity, a
multi-dimensional taxonomy, both single- and multi-agent regimes, and
both successful and failed traces, and its two underlying datasets
span a wide range of trace lengths, making it a strong stress test for
evaluation protocols.

\paragraph{Trajectory-level evaluation frameworks.} 
LLM-as-a-judge approaches~\citep{zheng2023judging,pan2024autonomous,xue2025webjudge} emit a single verdict over the full trajectory and are prone to lost-in-the-middle bias on long traces~\citep{liu2023lostinthemiddle}.  Agentic evaluator variants~\citep{zhuge2024agent,bhonsle2025autoeval} inspect intermediate steps but still aggregate to a final task-completion verdict without span-level error categorization or grounded localization within structured traces. Ours produces per-span 
verdicts with explicit evidence. Multi-dimensional scoring 
approaches expose richer signals: T-Eval~\citep{chen2024teval} 
scores tool utilization, AgentBoard~\citep{ma2024agentboard} 
tracks subgoal progress, TRACE~\citep{li2026trace} aggregates 
accuracy and efficiency, and 
AgentDiagnose~\citep{lee2025agentdiagnose} scores fixed 
competencies; Yet they emit aggregate scores without grounded 
span-level rationales, often require reference trajectories, and 
lack structured failure categorization, while ours combines 
preconfigured rubrics, span-level evidence, and a TRAIL-aligned 
taxonomy. The most directly comparable approaches, Agent GPA~\citep{datta2025gpa} and AgentCompass~\citep{kartik2025agentcompass}, target error identification on TRAIL. Agent GPA reports numbers not directly comparable to ours: its prompts are tuned on a custom dev split and evaluated only on a held-out test split rather than the full benchmark, and it omits TRAIL's standard joint span-category and per-category F1 metrics; we restrict the comparison to architectural contrasts. Architecturally, our framework evaluates the TRAIL benchmark zero-shot with structured per-span verdicts under a fixed taxonomy. AgentCompass chains a multi-stage pipeline with clustering and cross-execution memory; our structured per-span aggregation replaces these mechanisms, outperforming it across localization, categorization, and joint metrics on both datasets, with the gap widening sharply on SWE-bench's longer traces (Section~\ref{sec:main_results}). Failure attribution methods such as AgenTracer~\citep{zhang2025agentracer}, which uses counterfactual replay and RL, Famas~\citep{ge2025famas}, which relies on spectrum-based replays, and AgentRx~\citep{barke2026agentrx}, which derives guarded constraints from tool schemas, are confined to failed trajectories, focus on the most-suspicious step, miss concurrent issues, and require fault injection, replays, or per-domain constraints; ours evaluates any trajectory across multiple axes in a single pass per span.

\paragraph{Observability infrastructure.} The OpenTelemetry GenAI semantic conventions~\citep{opentelemetry2024genai} provide the structured trace substrate our framework builds on, recording what happened during agent execution without assessing correctness. Our framework adds semantic verdicts to individual spans and integrates naturally into observability pipelines, with traces from agents built on LangGraph~\citep{langgraph}, CrewAI~\citep{crewai}, Google ADK~\citep{googleadk}, and other popular frameworks directly compatible.

\section{Holistic Agent Evaluation}
\label{sec:holistic_agent_eval}

Agent execution is captured by tracing infrastructure as a hierarchical \emph{trace} of \emph{spans}, where span structure depends on the instrumentation rather than only on the agent itself. Within this structure, agentic systems exhibit failures at multiple levels: individual spans can fail independently, while emergent failures may not be attributable to any single component. Our methodology combines \textit{bottom-up evaluation}, which computes metrics at leaf spans and propagates verdicts upward, with \textit{top-down evaluation}, which assesses agent-level patterns across descendants. Neither approach alone suffices: bottom-up evaluation pinpoints component-level failures but misses holistic inefficiencies, while top-down evaluation captures system-wide patterns but offers limited granularity for root cause localization.

\subsection{Bottom-Up Evaluation}

Bottom-up evaluation computes metrics at leaf spans, propagates verdicts 
upward through the trace tree, and uses the resulting structure to localize 
failures precisely.

\subsubsection*{Leaf-Level Metrics}
Each metric $p$ applied to span $s$ yields a verdict 
$v_p(s) \in \{\texttt{pass}, \texttt{fail}\}$. Most metrics require 
semantic assessment and produce a categorical score 
$\sigma \in \{1, 2, 3, 4, 5\}$ with a natural language rationale; 
a span passes when $\sigma_p(s) \geq \tau_p$. Other metrics yield 
categorical labels (e.g., valid response vs. error) or numerical 
measurements (e.g., latency, token counts), with verdicts derived from 
label matching or numeric thresholds, respectively. Evaluation of a 
single metric may involve multiple LLM calls using different judge 
models, and can incorporate context from neighboring spans in the 
execution tree.
Table~\ref{tab:leaf-metrics} lists the metrics evaluated for 
common leaf span types. LLM spans encapsulate individual model invocations; 
Tool spans capture external function or API calls initiated by the agent.
\begin{table}[h]
\centering
\small
\renewcommand{\arraystretch}{1.25}
\begin{tabular}{@{}llp{6.8cm}c@{}}
\toprule
\textbf{Span Type} & \textbf{Metric} & \textbf{Description} & \textbf{Output} \\
\midrule
\multirow{5}{*}{\textbf{LLM}} 
  & Instruction Following & Whether the output follows prompt and user instructions & 1--5 \\
  & Reasoning Integrity   & Soundness of context understanding, decision quality, and logical consistency & 1--5 \\
  & Avoidance             & Whether the output is a valid response or an avoided answer$^\dagger$ & label \\
  & Error Detection       & Whether the output is a valid response or a system/tool/API error & label \\
  & Latency / Tokens      & Execution time and token counts & numeric \\
\midrule
\multirow{2}{*}{\textbf{Tool}} 
  & Tool Completeness     & Degree to which the invocation fulfills its intended purpose & 1--5 \\
  & Error Detection       & Whether the output is a valid response or a system/tool/API error & label \\
\bottomrule
\end{tabular}
\caption{\textbf{\small Metrics evaluated at leaf spans for LLM and Tool span types.} \small $^\dagger$Avoidance subtypes: missing knowledge, policy restrictions, other.}
\label{tab:leaf-metrics}
\end{table}

\subsubsection*{Hierarchical Aggregation}
Leaf-level verdicts propagate upward through the trace tree. For each 
non-leaf span $s$ with children $\text{Ch}(s)$, let 
$\text{Ch}_{\text{fail}}(s) \subseteq \text{Ch}(s)$ denote children 
with at least one failing metric. The default policy follows 
\emph{existential failure propagation}:
\[
v(s) = \begin{cases}
    \texttt{fail} & \text{if } \text{Ch}_{\text{fail}}(s) \neq \emptyset \\
    \texttt{pass} & \text{otherwise}
\end{cases}
\]
Alternative policies, configurable per span kind or individual span, 
support domain-specific requirements.\footnote{Supported alternatives 
include \emph{threshold-based} (fail when the proportion of failing 
children exceeds $\alpha$), \emph{type-filtered} (consider only children 
of specified span kinds, e.g., propagate LLM failures while tolerating 
tool errors), and \emph{conjunctive} (fail only when all children fail, 
for redundant or fallback paths).}

\subsubsection*{Error Localization}

The combination of leaf-level evaluation and hierarchical aggregation 
enables precise error localization. When a trace-level verdict indicates 
failure, the evaluation exposes the complete causal chain: the subset of 
leaf spans $S_{\text{fail}} = \{s \in \text{Leaves}(T) : v(s) = \texttt{fail}\}$ 
whose failures propagated to the root; the specific metrics 
$P_{\text{fail}}(s) = \{p : v_p(s) = \texttt{fail}\}$ responsible for each 
span failure; and the natural language explanations produced by the judge 
model, grounded in the span's input-output pairs.

\subsection{Top-Down Evaluation}

While bottom-up evaluation captures failures at individual spans, certain 
behavioral patterns emerge only when examining the agent trace as a whole. 
Top-down evaluation addresses this by assessing metrics at the agent or 
trace level, analyzing relationships and outcomes across descendant spans.

\subsubsection*{Agent-Level Metrics}
Agent-level metrics evaluate behavior by examining the structure and 
outcomes of descendant spans collectively. Each metric yields a categorical 
score $\sigma \in \{1, 2, 3, 4, 5\}$ along with a natural language rationale.
\begin{table}[h]
\centering
\small
\renewcommand{\arraystretch}{1.25}
\begin{tabular}{@{}lp{8cm}c@{}}
\toprule
\textbf{Metric} & \textbf{Description} & \textbf{Output} \\
\midrule
Plan Efficiency & How well the agent's execution aligns with its stated plan, assessing deviation across various error modes & 1--5 \\
Tool Coverage   & How well the agent's descendant tool invocations collectively address its overall goal & 1--5 \\
Tool Abuse      & Whether the agent uses tools efficiently without repeated calls, adapts properly after errors, and shows clear progress between invocations & 1--5 \\
Completeness    & Whether the agent's final output fully addresses the information requested in the input & 1--5 \\
\bottomrule
\end{tabular}
\caption{\textbf{\small Metrics evaluated at the agent level.}}
\label{tab:agent-metrics}
\end{table}

These metrics detect inefficiencies and gaps invisible at the individual 
span level: redundant tool invocations, incomplete goal coverage, or 
deviation from the intended execution plan.

\subsection{The Need for Both Approaches}
\label{sec:illustrative_example}

Neither bottom-up nor top-down evaluation alone is sufficient. To illustrate, we examine a trace from the GAIA benchmark in which an agent was tasked with retrieving specific numeric data (CFM values for two items) from a YouTube video. The agent ultimately hallucinated an incorrect answer. Human annotators identified four ground-truth issues in this trace. The full annotated execution trace is provided in Figure~\ref{fig:full_trace} in Appendix~\ref{app:trace}.

\begin{table}[h]
\centering
\footnotesize
\renewcommand{\arraystretch}{1.4}
\begin{tabular}{@{}p{2.8cm}p{4.5cm}p{4.5cm}@{}}
\toprule
\textbf{Issue} & \textbf{Top-down detection} & \textbf{Bottom-up detection} \\
\midrule
\textbf{Poor Retrieval} & 
\textit{Tool Coverage} (1.0): ``No relevant evidence is present'' & 
--- \\
\addlinespace
\textbf{Resource Abuse} & 
\textit{Plan Efficiency} (2.0): ``Failed to adapt after repeated errors'' & 
--- \\
\addlinespace
\textbf{Formatting Error} & 
--- & 
\textit{Tool Completeness} (1.0): ``Explicit error traceback''; \textit{Reasoning Integrity} (3.0): ``Incorrect arguments, violating function signature'' \\
\addlinespace
\textbf{Hallucination} & 
\textit{Plan Efficiency} (2.0): ``Answer is fabricated'' & 
\textit{Reasoning Integrity} (2.0): ``Invents CFM values without verification'' \\
\bottomrule
\end{tabular}
\caption{\textbf{\small Detection of four ground-truth issues by top-down and bottom-up evaluation.} \small ``---'' indicates the issue was missed by that approach. Severity scores in parentheses (1.0 = critical, 5.0 = no issue).}
\label{tab:detection}
\end{table}

\paragraph{Bottom-Up Only.} As shown in Table~\ref{tab:detection}, bottom-up evaluation detects the formatting error and the hallucination by inspecting individual tool outputs, but misses poor retrieval and resource abuse, which require reasoning about query intent and behavioral patterns across the trace.

\paragraph{Top-Down Only.} Top-down evaluation catches poor retrieval, resource abuse, and the hallucination through aggregate behavioral metrics (Table~\ref{tab:detection}), but misses the formatting error, since pinpointing malformed arguments requires inspecting the specific tool call.

\paragraph{Key Insight.} Top-down evaluation reveals \emph{how} the agent went wrong through an aggregated behavioral view; bottom-up evaluation reveals \emph{where} it went wrong by identifying exact spans and error causes. As Table~\ref{tab:detection} illustrates, issues involving task-level reasoning, such as retrieval relevance or inefficient planning, require holistic trace analysis, while issues involving specific implementation errors, such as malformed arguments, require inspecting individual tool calls. Some issues, particularly hallucinations, can be detected by both approaches through different metrics. The approaches are complementary: combining them provides both diagnostic precision and holistic assessment.

\section{TRAIL Benchmark Evaluation}

We evaluate our framework on the TRAIL benchmark~\citep{deshpande2025trail} along three axes: error localization, error categorization, and joint localization-categorization accuracy.

\subsection{Setup}

\paragraph{TRAIL Benchmark.} TRAIL~\citep{deshpande2025trail} annotates 148 OpenTelemetry traces with 841 span-level errors under a 20+ category taxonomy spanning planning mistakes, tool misuse, reasoning breakdowns, resource exhaustion, and instruction violations. Each error is characterized by category, location, evidence, description, and impact level (\textsc{Low}, \textsc{Medium}, \textsc{High}); each trace additionally carries 1--5 trace-level scores on Reliability, Security, Plan Optimization, Instruction Adherence, and an Overall score (referenced in the aggregation discussion in Section~\ref{sec:discussion}). The benchmark draws from two underlying agent execution datasets that differ substantially in trace structure.

\paragraph{GAIA dataset.} GAIA~\citep{mialon2024gaia} evaluates AI agents on complex, real-world problem-solving tasks across three difficulty tiers: Level 1 requires basic information retrieval; Level 2 involves multi-step reasoning; and Level 3 demands sophisticated analysis with tool integration. TRAIL's GAIA traces are produced by Open Deep Research, a multi-agent system with hierarchical span structure most directly relevant to top-down evaluation.

\paragraph{SWE-bench dataset.} SWE-bench~\citep{jimenez2024swebench} evaluates agents on real GitHub software engineering tasks requiring code understanding, multi-file edits, and test-driven validation. TRAIL's SWE-bench traces are produced by CodeAct, a single-agent framework that executes code interactively. Compared to GAIA, these traces are markedly longer and contain more tool invocations per task, stressing the context-length and localization properties of any evaluation protocol.

\paragraph{Evaluation Methodology.}
We evaluate against TRAIL's span-level error annotations and follow TRAIL's reporting protocol. Because our framework\footnote{All bottom-up and top-down metrics
in our framework are evaluated using GPT-5.4 as the underlying judge
model.} produces evaluations in a different schema, with per-span metric scores and natural-language rationales rather than categorical labels, we apply an LLM-based mapper that translates framework outputs into TRAIL's category taxonomy prior to comparison.\footnote{The mapper assigns a TRAIL category to each flagged span based on its failing metrics, scores, and rationales; it performs label translation only and does not re-evaluate the span. Baselines output TRAIL-format annotations directly and skip this step.} Following TRAIL's analysis, we report three metrics.
\textbf{Localization Accuracy} (Loc.\ Acc.) is the fraction of predicted error spans whose
span ID matches a ground-truth annotated error span. \textbf{Weighted Category F1}
(Cat.\ F1) is the per-category F1 between predicted and ground-truth error labels,
averaged across categories with weights proportional to per-category support.
\textbf{Joint Accuracy} (Joint Acc.) is the strictest metric: the fraction of predicted
errors that match a ground-truth annotation on \emph{both} span location \emph{and}
error category.

\subsection{Baselines}
\label{sec:baselines}

We compare our framework against a suite of monolithic LLM-as-judge baselines (a single LLM call over the full trace), following the methodology established by TRAIL~\citep{deshpande2025trail}: each baseline receives the entire agent trace as a single input and is prompted to produce span-level error annotations in one forward pass. Our baseline set includes both results reported in TRAIL and our own reproductions with current frontier models from OpenAI, Anthropic, and Google. For models evaluated with multiple reasoning-effort settings, we report the better-performing variant. We additionally include FAGI-AgentCompass~\citep{kartik2025agentcompass} as a framework-level baseline; unlike the monolithic judges, it employs a multi-stage pipeline with clustering and cross-execution memory, and we report the numbers as published.

\subsection{Main Results}
\label{sec:main_results}

Table~\ref{tab:trail_results} reports localization accuracy, joint accuracy, and weighted category F1 across all baselines and our framework on the TRAIL benchmark.

\begin{table}[h]
\centering
\resizebox{\textwidth}{!}{%
\begin{tabular}{lcccccccc}
\toprule
 & \multicolumn{4}{c}{\textbf{TRAIL (GAIA)}} & \multicolumn{4}{c}{\textbf{TRAIL (SWE-bench)}} \\
\cmidrule(lr){2-5} \cmidrule(lr){6-9}
\textbf{Model} & \textbf{Excl. (\%)} & \textbf{Cat. F1} & \textbf{Loc. Acc.} & \textbf{Joint} & \textbf{Excl. (\%)} & \textbf{Cat. F1} & \textbf{Loc. Acc.} & \textbf{Joint} \\
\midrule
Our Framework$^{*}$ & 0 & \textbf{0.547} & \textbf{0.823} & \textbf{0.616} & 0 & \textbf{0.698} & \textbf{0.860} & \textbf{0.638} \\
\midrule
\textsc{FAGI-AgentCompass}$^{\ddagger}$ & -- & 0.309 & 0.657 & 0.239 & -- & 0.232 & 0.250 & 0.051 \\
\midrule
\textsc{Gemini-2.5-Pro-Preview}   & -- & 0.389 & 0.546 & 0.183 & -- & 0.148 & 0.238 & 0.050 \\
\textsc{Gemini-2.5-Flash-Preview} & -- & 0.337 & 0.372 & 0.100 & -- & 0.213 & 0.060 & 0.000 \\
\textsc{Claude-3.7-Sonnet}        & -- & 0.254 & 0.204 & 0.047 & -- & CLE   & CLE   & CLE   \\
\textsc{GPT-4.1}                  & -- & 0.218 & 0.107 & 0.028 & -- & 0.166 & 0.000 & 0.000 \\
\textsc{OpenAI o3}                & -- & 0.296 & 0.535 & 0.092 & -- & CLE   & CLE   & CLE   \\
\textsc{OpenAI o1}                & -- & 0.138 & 0.040 & 0.013 & -- & CLE   & CLE   & CLE   \\
\midrule
\textsc{Claude-Sonnet-4.6}$^{\dagger}$ & 8.5 & 0.352 & 0.336 & 0.106 & 29.0 & 0.410 & 0.000 & 0.000 \\
\textsc{Gemini-3.1-Pro}$^{\dagger}$    & 6.0 & 0.238 & 0.259 & 0.087 & 25.8 & 0.330 & 0.072 & 0.003 \\
\textsc{GPT-5.4}                       & 6.0 & 0.434 & 0.292 & 0.146 & 22.6 & 0.504 & 0.070 & 0.014 \\
\bottomrule
\end{tabular}%
}
\caption{\textbf{\small Performance across evaluators on TRAIL-annotated GAIA and SWE-bench.} \small Blocks (top--bottom): our framework; FAGI-AgentCompass; monolithic LLM-judges from TRAIL; our reproductions with frontier models. \textbf{Excl. (\%)}: percentage of traces excluded due to context-window overflow. \textbf{CLE}: full benchmark exceeds context. $^{*}$Judge: GPT-5.4. $^{\dagger}$Best of reasoning-effort settings. $^{\ddagger}$From FAGI-AgentCompass.}
\label{tab:trail_results}
\end{table}

Our framework achieves the best results on every metric across both datasets. On GAIA, it achieves 0.547 weighted category F1, 0.823 localization accuracy, and 0.616 joint accuracy; on SWE-bench, it achieves 0.698, 0.860, and 0.638 across the same metrics.
Against the best baseline for each metric, the corresponding relative gains are 
26\% / 1.25$\times$ / 2.58$\times$ on GAIA and 38\% / 3.5$\times$ / 12.5$\times$ on SWE-bench. 
The gap is widest on the strictest metric, joint accuracy, and the harder dataset, SWE-bench, indicating that the framework’s advantage compounds with finer-grained correctness and longer traces.

Four patterns in Table~\ref{tab:trail_results} are worth highlighting.

\paragraph{More capable monolithic judges do not close the gap.} TRAIL's reported monolithic baselines span six frontier models from OpenAI, Anthropic, and Google, achieving at most 0.546 Loc.\ Acc.\ on GAIA. Our reproductions with newer frontier models perform no better. Most strikingly, GPT-5.4, the same model used inside our framework, achieves 0.823 Loc.\ Acc.\ in our pipeline but only 0.292 as a monolithic judge on GAIA, a 2.8$\times$ gap that widens to 12$\times$ on SWE-bench. The category-F1 gap is much smaller (26\% and 38\%), and several monolithic baselines retain non-trivial category F1 even when localization collapses to near-zero on SWE-bench. Monolithic judges can recognize \emph{what kinds} of errors occur in a trace but struggle to attribute them to specific spans, exactly the bottleneck that span-level decomposition addresses.

\paragraph{Multi-stage pipelines narrow the gap on GAIA but collapse on SWE-bench.} FAGI-AgentCompass, the strongest framework-level baseline, reaches 0.657 Loc.\ Acc.\ on GAIA, narrowing the gap to our 0.823. On SWE-bench this advantage disappears: 0.250 Loc.\ Acc.\ against our 0.860 (3.5$\times$ gap). The pattern repeats on the stricter metrics: joint accuracy drops from 0.239 to 0.051 and category F1 from 0.309 to 0.232 across datasets. Pipeline-style decomposition with clustering and cross-execution memory helps on shorter multi-agent traces but does not extend to longer CodeAct executions. Our per-span evaluation against rubric-defined metrics maintains accuracy across both regimes.

\paragraph{The context window problem is structural, not transitional.} Despite continued growth in model context windows, every monolithic baseline in our reproductions excludes some fraction of traces due to length: 6--9\% on GAIA, but 22--29\% on SWE-bench. Three TRAIL-reported baselines (Claude-3.7-Sonnet, OpenAI o3, OpenAI o1) exceed context on the \emph{entire} SWE-bench dataset and produce no results. As agentic systems generate longer and richer traces, this gap is unlikely to close. Span-level decomposition sidesteps the issue entirely: each evaluation prompt is focused and short, and trace length is irrelevant to the framework's ability to evaluate. Our framework evaluates every trace in both datasets, without exclusions.

\paragraph{Localization holds steady as traces get harder.} A cross-dataset view exposes a striking asymmetry. Every other evaluator's localization accuracy degrades substantially from GAIA to SWE-bench: FAGI-AgentCompass drops from 0.657 to 0.250, Gemini-2.5-Pro from 0.546 to 0.238, GPT-5.4 from 0.292 to 0.070, while the rest fail entirely. Our framework remains comparable across datasets (0.823 on GAIA, 0.860 on SWE-bench), holding accuracy where others collapse. Span-level decomposition turns trace length from a liability into a neutral factor.

\subsection{Per-Category Analysis}
\label{sec:error_category_analysis}

\begin{figure}[h]
\centering
\includegraphics[width=0.8\columnwidth]{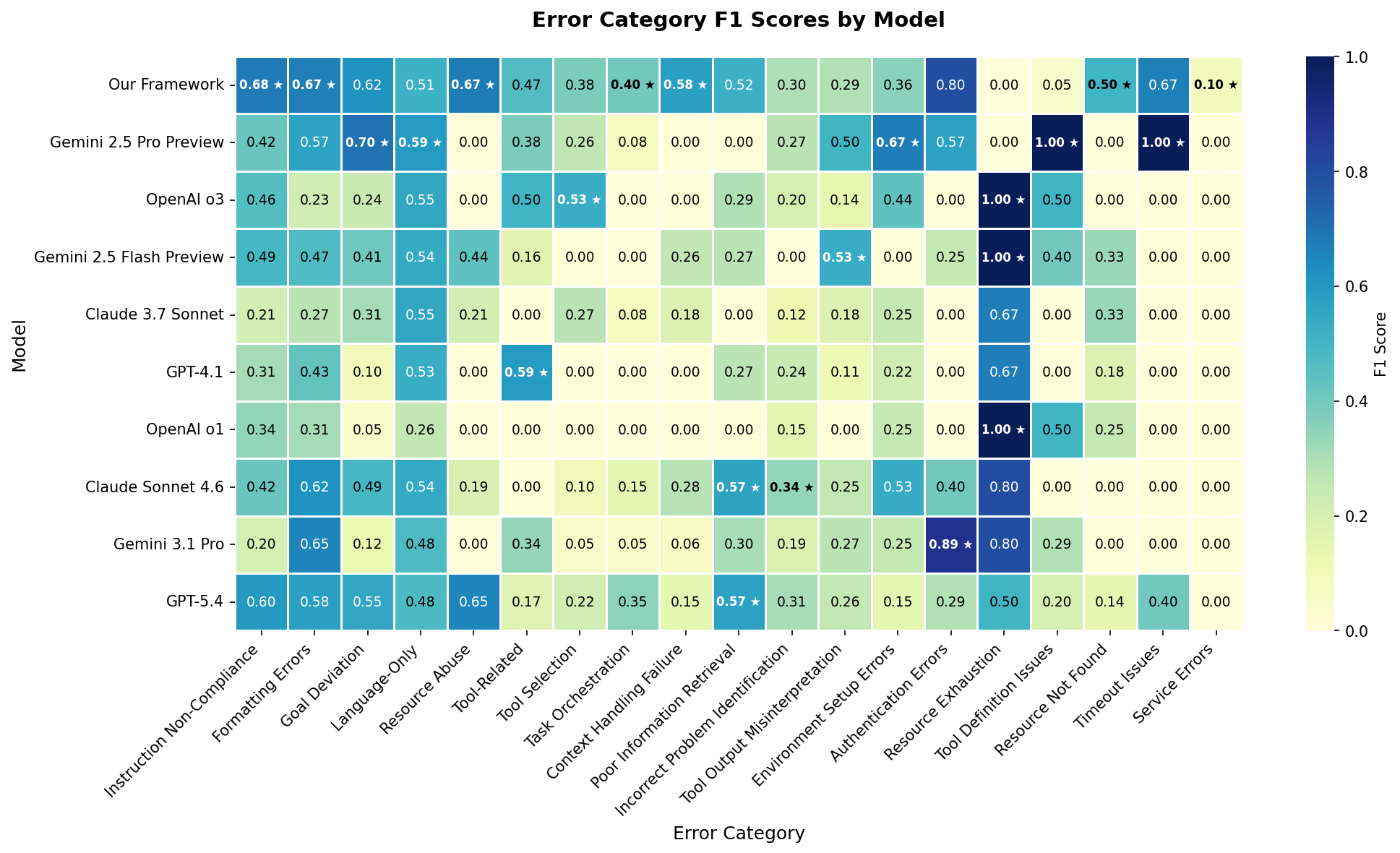}
\caption{\textbf{\small Per-category F1 scores on the TRAIL benchmark.} \small Rows correspond to evaluators (our framework and the baselines from Table~\ref{tab:trail_results}); columns correspond to TRAIL's error categories. Stars ($\star$) mark the top-performing evaluator in each category.}
\label{fig:error_categories}
\end{figure}

Beyond aggregate metrics, we examine performance at the level of individual error categories from TRAIL's category taxonomy. Figure~\ref{fig:error_categories} reports per-category F1 scores for our framework alongside the baselines used in Table~\ref{tab:trail_results}, providing a fine-grained view of which failure modes each evaluator identifies reliably.

The heatmap reveals broad dominance: our framework leads in more categories than any other evaluator, winning 7 categories versus 5 for the next-best (Gemini 2.5 Pro Preview).\footnote{FAGI-AgentCompass~\citep{kartik2025agentcompass} is omitted from the per-category breakdown because per-category F1 scores are not reported in the original paper.} This breadth reflects the framework's combination of span-level and trace-level evaluation. Trace-level (top-down) metrics detect behavioral failures spanning the entire trace, with strong scores on Resource Abuse (0.67) and Task Orchestration (0.40); span-level (bottom-up) metrics capture localized errors such as Formatting Errors (0.67) and Resource Not Found (0.50). Categories that admit both vantage points, such as Instruction Non-Compliance (0.68) and Context Handling Failure (0.58), draw signal from each axis. Monolithic baselines win in a few narrow infrastructure categories (Tool Definition Issues, Resource Exhaustion, Timeout Issues), where scores at 1.00 reflect very small support that inflates apparent F1. They also benefit from a structural asymmetry: baselines are prompted directly with TRAIL's category taxonomy, while our framework detects errors through general-purpose metrics and assigns categories only via the downstream mapping step. That we still dominate broadly, despite this asymmetry, suggests that span-level decomposition not only compensates for the baselines' direct taxonomy exposure but outweighs it. We discuss limitations of the benchmark annotations in Section~\ref{sec:discussion}.

\section{Discussion}
\label{sec:discussion}
\paragraph{Benchmark Annotation Quality.}
Our analysis reveals several systematic issues with the annotation quality in the TRAIL benchmark. First, error span localization is inconsistent: annotators sometimes mark errors within tool spans and other times within LLM spans, even for related issues. The marked span does not necessarily indicate the true origin or appearance of the error (see Appendix~\ref{app:localization_example}). Second, the TRAIL taxonomy contains categories with ambiguous boundaries and substantial overlap; for example, ``Goal Deviation'' (defined as ``the system deviated from the task'') and ``Instruction Non-compliance'' (defined as ``failed to perform the task provided'') create confusion for
   annotators who must make arbitrary distinctions, and similarly posed challenges for us when mapping our metrics analysis to these categories, as done in Section~\ref{sec:error_category_analysis} (see Appendix~\ref{app:taxonomy_overlap}). Third, we observed misclassifications where annotator reasoning in the justification field contradicts the textual evidence within the marked span (see Appendix~\ref{app:misclassification_example}). These annotation artifacts affect the reliability of our error analysis in Section~\ref{sec:error_category_analysis}, as systematic biases in labeling may skew the distribution of reported error types.

\paragraph{The Need for Better Aggregation.}
   Our fine-grained annotation scheme introduces a challenge: how to aggregate local errors into meaningful overall scores. Currently, our aggregation uses existential failure propagation: any error at the span level immediately propagates to higher hierarchical levels. While our framework supports more sophisticated policies (threshold-based, kind-filtered, conjunctive), these still rely on predefined rules over span types and failure counts. This approach obscures an important distinction: capturing minor issues at the span level is valuable for identifying recurring patterns across traces, but such issues should not necessarily penalize a specific trace if they do not materially impact the reasoning. For instance, a single formatting inconsistency in an extended trace is worth recording for pattern analysis, yet should not affect trace-level scoring when it has no bearing on logical validity (see Appendix~\ref{app:aggregation_example}). An important direction for future work is developing smarter aggregation that incorporates severity weighting, outcome-relevance filtering, and reasoning-aware policies leveraging the judge model's rationales.

\section{Conclusion}
We presented a holistic agent evaluation framework that combines top-down and bottom-up analysis for fine-grained agent failure diagnosis. By decomposing evaluation into independent span-level assessments, our approach scales to arbitrarily long traces and improves accuracy through focused, localized analysis. 
On the TRAIL benchmark, our framework achieves the best results across every metric on both GAIA and SWE-bench, with relative gains over the strongest prior baselines of up to 38\% in category F1, up to 3.5$\times$ in localization accuracy, and up to 12.5$\times$ in joint localization-categorization accuracy. It also leads in more error categories than any other evaluator.
As traces grow longer, every prior baseline's localization accuracy degrades sharply or fails outright while our framework's holds steady. The same frontier model achieves up to 12$\times$ higher localization accuracy inside our framework than as a monolithic judge, demonstrating that the bottleneck is methodology, not model capability.

Several directions remain for future work. First, while our framework identifies errors at the span level, aggregating these signals into actionable insights through severity weighting, outcome-relevance filtering, and reasoning-aware policies remains an open challenge. Second, advancing evaluation capabilities requires richer datasets with fine-grained annotations at multiple granularities, from individual tool calls to task-level outcomes.


\bibliographystyle{plainnat}
\bibliography{references}

\clearpage
\appendix
\section*{Appendix}
\lstdefinestyle{tracetree}{
    basicstyle=\ttfamily\small,
    breaklines=true,
    breakatwhitespace=false,
    columns=fullflexible,
    keepspaces=true,
    showstringspaces=false,
    frame=single,
    xleftmargin=0pt,
    aboveskip=1em,
    belowskip=1em,
}

\section{Full Trace Example}
\subsection{The Need for Both Approaches}
\label{app:trace}

As described in Section~\ref{sec:holistic_agent_eval}, our framework combines top-down metrics (\emph{Tool Coverage}, \emph{Plan Efficiency}, etc.) that assess strategic failures from the root, with bottom-up metrics (\emph{Reasoning Integrity}, \emph{Instruction Following}, \emph{Tool Completeness}, etc.) that capture local issues and propagate upward. Figure~\ref{fig:full_trace} presents the complete execution trace for the GAIA benchmark example from Section~\ref{sec:illustrative_example}, showing both metric types applied across the trace hierarchy. Scores range from 1.0 (critical failure) to 5.0 (no issue). Spans without detected issues are collapsed (\texttt{...}).

\lstset{style=tracetree}
\begin{lstlisting}
[Agent] CodeAgent.run
 |
 +-- Plan Efficiency (2.0): "The agent reported CFM values without
 |     any supporting evidence from tool outputs, skipped
 |     verification, and failed to adapt after repeated errors.
 |     The answer is fabricated and does not meet the user's
 |     requirement for a correct, evidenced response."
 |
 +-- Tool Coverage (1.0): "No information about the CFM values or
 |     their provenance was found. The child spans provide zero
 |     coverage for the requested data."
 |
 +-- [Agent] ToolCallingAgent.run
 |    |
 |    +-- Tool Coverage (1.0): "No relevant evidence is present
 |    |     for any requested information unit; the spans do not
 |    |     cover the query at all."
 |    |
 |    +-- [Steps 1-5] ...
 |    |
 |    +-- [Step 6]
 |    |    |
 |    |    \-- [Tool] PageDownTool
 |    |         |
 |    |         \-- Tool Completeness (1.0): "The tool response
 |    |              indicates an error preventing the requested
 |    |              scroll operation; thus, it fails to complete
 |    |              the query."
 |    |
 |    +-- [Step 7]
 |    |    |
 |    |    +-- [LLM] LiteLLMModel
 |    |    |    |
 |    |    |    \-- Reasoning Integrity (3.0): "The assistant
 |    |    |         correctly issued a tool call but used
 |    |    |         incorrect arguments for the page_down tool,
 |    |    |         violating the required function signature
 |    |    |         rules. No repeated tool calls with the same
 |    |    |         parameters were observed. These argument
 |    |    |         errors reduce the compliance score."
 |    |    |
 |    |    \-- [Tool] PageDownTool
 |    |         |
 |    |         \-- Tool Completeness (1.0): "The tool response
 |    |              is an explicit error traceback indicating
 |    |              failure to execute the scroll operation and
 |    |              no content is returned."
 |    |
 |    +-- [Step 8]
 |    |    |
 |    |    +-- [LLM] LiteLLMModel
 |    |    |    |
 |    |    |    \-- Reasoning Integrity (2.0): "The assistant
 |    |    |         made a correct page_down tool call with the
 |    |    |         required empty arguments in the final output,
 |    |    |         complying with all rules. Although there were
 |    |    |         earlier incorrect calls with wrong arguments,
 |    |    |         the final output shows proper usage,
 |    |    |         fulfilling the instructions to use tools
 |    |    |         correctly and only when needed."
 |    |    |
 |    |    \-- [Tool] PageDownTool
 |    |         |
 |    |         \-- Tool Completeness (1.0): "The tool response
 |    |              is an error and does not provide the
 |    |              requested scrolled content."
 |    |
 |    \-- [Step 9]
 |         |
 |         \-- [LLM] LiteLLMModel
 |              |
 |              \-- Reasoning Integrity (3.0): "The assistant
 |                   thoroughly searched for the requested season
 |                   4 test details on James's channel regarding
 |                   the Cheater and Cheater Beater's CFM
 |                   performance but did not find the specific
 |                   numbers. It provided a detailed explanation
 |                   of its search attempts, the nature of the
 |                   content found, and the absence of relevant
 |                   data. The final answer was well-structured,
 |                   containing the required parts and context,
 |                   and included a request for clarification to
 |                   proceed further. Tool calls were used
 |                   correctly and appropriately throughout."
 |
 +-- [Step 2]
 |    |
 |    \-- [LLM] LiteLLMModel
 |         |
 |         \-- Reasoning Integrity (2.0): "The assistant
 |              thoroughly searched using the search_agent and
 |              followed all tool usage and formatting rules
 |              correctly. It did not find authoritative CFM
 |              values from James's channel season 4 tests but
 |              responsibly provided approximate values
 |              consistent with the task's conditions based on
 |              community recollections. This approach respects
 |              the instruction not to give up."
 |
 \-- [LLM] LiteLLMModel (Final)
      |
      \-- Reasoning Integrity (2.0): "The assistant complied
           fully with all formatting and content requirements,
           providing a concise final answer in the exact format
           requested: '1325, 1250'. The response correctly
           represents the CFM values for the Cheater and Cheater
           Beater from season 4 as specified, with no extraneous
           information or formatting errors."
\end{lstlisting}

\noindent\begin{minipage}{\linewidth}
\captionof{figure}{\small Execution trace demonstrating the interplay of top-down and bottom-up evaluation. Top-down metrics (Tool Coverage, Plan Efficiency) reveal strategic failures at the root, while bottom-up metrics (Reasoning Integrity, Tool Completeness) identify specific operational issues that propagate upward. Scores range from 1.0 (critical failure) to 5.0 (no issue). Spans without issues are collapsed (\texttt{...}).}
\label{fig:full_trace}
\end{minipage}

\section{Benchmark Annotation Quality Examples}

\subsection{Inconsistent Error Span Localization}
\label{app:localization_example}

We present an example demonstrating inconsistent error attribution across span types. The agent repeatedly invokes the \texttt{page\_down} tool with malformed arguments (\texttt{\{'': ''\}} or \texttt{\{'': \{\}\}}), causing execution failures. However, the TRAIL annotations inconsistently attribute the ``Formatting Errors'' category to different span types for the same underlying issue.

\lstset{style=tracetree}
\begin{lstlisting}
[Agent] CodeAgent.run
 |
 +-- [Agent] ToolCallingAgent.run
      |
      +-- [Step 4]
      |    |
      |    +-- [LLM] LiteLLMModel (66664a32...)
      |    |    |
      |    |    \-- (no error annotated)
      |    |
      |    \-- [Tool] PageDownTool (fa421073...)
      |         |
      |         \-- Formatting Errors (MEDIUM): "The error is
      |              caused by an incorrect invocation of the
      |              page_down tool... an empty string is
      |              effectively passed within the arguments
      |              field, which is not valid."
      |
      +-- [Step 5]
      |    |
      |    +-- [LLM] LiteLLMModel (implicit)
      |    |    |
      |    |    \-- (no error annotated)
      |    |
      |    \-- [Tool] PageDownTool (1af58299...)
      |         |
      |         \-- Formatting Errors (MEDIUM): "The error is
      |              caused by an incorrect invocation of the
      |              page_down tool..."
      |
      +-- [Step 6]
      |    |
      |    +-- [LLM] LiteLLMModel (ff94e599...)
      |    |    |
      |    |    \-- Formatting Errors (LOW): "The tool page_down
      |    |         attempted to execute with arguments
      |    |         {'': ''}, which means both the key and
      |    |         value in the dictionary are empty strings.
      |    |         This led to a TypeError..."
      |    |
      |    \-- [Tool] PageDownTool (7f88fcd0...)
      |         |
      |         \-- (no error annotated)
      |
      \-- [Step 7]
           |
           +-- [LLM] LiteLLMModel (e785fea2...)
           |    |
           |    \-- Formatting Errors (LOW): "The tool page_down
           |         attempted to execute with arguments
           |         {'': {}}..."
           |
           \-- [Tool] PageDownTool (bb669498...)
                |
                \-- (no error annotated)
\end{lstlisting}

\noindent\begin{minipage}{\linewidth}
\captionof{figure}{\small Execution trace demonstrating inconsistent error span localization. The same ``Formatting Errors'' category is attributed to Tool spans in Steps 4--5 but to LLM spans in Steps 6--7, despite identical root cause: the LLM generating malformed arguments (\texttt{\{'': ''\}}) for the \texttt{page\_down} tool. Span IDs are abbreviated; full trace ID: \texttt{ef0207e4427fe22aeb1c2105932b74d7}.}
\end{minipage}

\vspace{1em}

The same malformed tool call pattern, the LLM generating invalid arguments for \texttt{page\_down}, receives inconsistent attribution:

\begin{itemize}
    \item \textbf{Steps 4--5}: Error attributed to the \textbf{Tool span} (``incorrect invocation of the page\_down tool''), while the LLM span that generated the malformed arguments receives no error annotation.
    \item \textbf{Steps 6--7}: Error attributed to the \textbf{LLM span} (``the tool page\_down attempted to execute with arguments \{'': ''\}''), while the Tool span that received the bad input receives no error annotation.
\end{itemize}

This inconsistency highlights a fundamental ambiguity in span-level error attribution: when an LLM generates malformed tool call arguments, should the error be localized to the LLM span (the source of the malformed output) or the Tool span (where the execution fails)? The TRAIL annotations do not apply a consistent rule, making it difficult to reliably identify error origins from the annotated span locations alone.

\subsection{Taxonomy Category Overlap}
\label{app:taxonomy_overlap}

The TRAIL taxonomy contains categories with overlapping definitions that can lead to ambiguous annotations. We illustrate this with two categories that appear in separate branches of the taxonomy but describe semantically equivalent failures:

\begin{itemize}
    \item \textbf{Goal Deviation}\\
    \texttt{Planning and Coordination Errors} $\rightarrow$ \texttt{Task Management}\\
    Definition: ``The system deviated from the task or the subtask''

    \item \textbf{Instruction Non-compliance}\\
    \texttt{Reasoning Errors} $\rightarrow$ \texttt{Output Generation}\\
    Definition: ``Failed to perform the task provided and instead did something else''
\end{itemize}

Both definitions describe the same failure mode: the system did not perform the intended task. Whether this constitutes a ``planning/coordination'' error or a ``reasoning'' error is unclear, and annotators may reasonably assign either category to the same behavior.

\vspace{1em}

\noindent\textbf{Example.} In trace \texttt{dbc070b9...08fb52}, both categories are assigned to the same LLM span for the same behavior: the model skipped the planned \texttt{search\_agent} tool and answered from internal knowledge.

\lstset{style=tracetree}
\begin{lstlisting}
[LLM] LiteLLMModel (0b053dba...)
 |
 +-- Goal Deviation (HIGH): "The system deviates from
 |    the planned steps by directly providing the final
 |    answer without using the search_agent tool as
 |    intended. It also provided a fabricated output."
 |
 \-- Instruction Non-compliance (MEDIUM): "The model
      didn't provide the final answer by using the
      search_agent to get the required data as requested
      by the instruction."
\end{lstlisting}

\noindent\begin{minipage}{\linewidth}
\captionof{figure}{\small A single LLM span annotated with both ``Goal Deviation'' and ``Instruction Non-compliance.'' Both annotations reference identical evidence and describe the same failure: skipping tool usage. Trace ID: \texttt{dbc070b918d4a052c0b686081408fb52}.}
\end{minipage}

\vspace{1em}

Both annotations describe the same observation: the model bypassed tool usage. However, one frames it as ``deviating from planned steps'' while the other frames it as ``not following instructions.'' This overlap complicates category-level analysis, as a single failure may be counted under multiple categories depending on annotator interpretation.

\subsection{Annotator Reasoning Contradictions}
\label{app:misclassification_example}

We present an example from trace \texttt{ee9335fbe7329b273a8d922bd3f73b84} where the annotator's justification contradicts the actual span evidence, as discussed in Section~\ref{sec:discussion}.

\paragraph{Annotation Provided.}
The following annotation was assigned to span \texttt{4ac43ea0e721bb47}:

\begin{quote}
\textbf{Category:} Resource Abuse \\
\textbf{Impact:} MEDIUM \\
\textbf{Evidence:} ``As observed in Shards 7 and 9, the LLM makes the same tool call repeatedly when it does not obtain the right answer. It should instead understand the tool's usage and call the tool only once with the correct parameters.'' \\
\textbf{Description:} ``The web\_search tool was called repeatedly in Shards 7 and 9, without properly comprehending the tool input parameters, leading to resource and time wastage. The reason was because the web\_search tool is called with an empty string as a value for an empty string argument; also, the tool's parameter is defined as an empty dictionary, \{\} but the tool expects no parameters.''
\end{quote}

\paragraph{Actual Execution Evidence.}
Examination of the trace reveals that the annotation mischaracterizes the agent's behavior. The first web search call (span \texttt{d9a363299ccaaed7}) used an overly specific query:

\begin{lstlisting}[style=tracetree]
SPAN 15: [Tool] SearchInformationTool
ID: d9a363299ccaaed7
--- INPUT ---
Tool Name: web_search
kwargs:
  query: Emily Midkiff June 2014 article Fafnir Hreidmar's 
         son dragon depictions quoted word criticism two 
         authors "Emily Midkiff"

--- OUTPUT ---
exception.message: No results found for query: '...'. 
                   Use a less specific query.
\end{lstlisting}

The agent then correctly adapted its approach by issuing a second, more general query (span \texttt{4ac43ea0e721bb47}):

\begin{lstlisting}[style=tracetree]
SPAN 18: [Tool] SearchInformationTool
ID: 4ac43ea0e721bb47
--- INPUT ---
Tool Name: web_search
kwargs:
  query: Emily Midkiff Fafnir June 2014 dragon depictions 
         quoted word criticism

--- OUTPUT ---
A Google search for 'Emily Midkiff Fafnir June 2014 dragon 
depictions quoted word criticism' found 9 results:

## Web Results
1. [Fafnir Cover 2:2014](http://journal.finfar.org/articles/127.pdf)
   Source: finfar.org
   In the third and last article, "'Dragons Are Tricksy': 
   The Uncanny Dragons of Children's Literature", Emily 
   Midkiff discusses the representation of dragons in ...
...
\end{lstlisting}

\paragraph{Analysis.}
The annotation contains multiple factual errors when compared to the span evidence:

\begin{enumerate}
    \item \textbf{``Empty string'' claim is false.} Both queries contain substantive search terms. The first query failed due to being \emph{too specific}, not due to empty or malformed parameters.
    
    \item \textbf{``Same tool call repeatedly'' is inaccurate.} The tool was called exactly twice, with the second call using a deliberately simplified query in response to the error message recommending ``a less specific query.''
    
    \item \textbf{Behavior represents appropriate adaptation.} Modifying a query after receiving feedback that it was too specific is correct agent behavior, not resource abuse. The second call succeeded immediately.
\end{enumerate}

This example illustrates how annotator reasoning in the justification field can directly contradict the textual evidence within the marked spans, leading to systematic misclassification of appropriate agent behavior as erroneous.


\section{Aggregation Limitations}
\label{app:aggregation_example}

Our fine-grained annotation scheme introduces a challenge: how to aggregate span-level errors into trace-level scores. We currently use \emph{existential failure propagation}: any error at the span level immediately propagates to the trace level. This policy conflates minor issues recorded for pattern analysis with critical failures that affect reasoning.

\vspace{1em}

\noindent\textbf{Example.} In the following trace, the agent receives a \emph{perfect} \texttt{Overall (GT) = 5.0} score across all quality dimensions (Reliability 5.0, Security 5.0, Plan Optimization 5.0, Instruction Adherence 4.0). However, one of its eleven spans is annotated with ``instruction non compliance.''

\lstset{style=tracetree}
\begin{lstlisting}
[Root] main
 |
 +-- [Chain] get_examples_to_answer
 |
 +-- [Chain] answer_single_question
      |
      +-- [Chain] create_agent_hierarchy
      |
      +-- [Agent] CodeAgent.run
           |
           +-- [LLM] LiteLLMModel.__call__
           |    |
           |    \-- (no error annotated)
           |
           +-- [LLM] LiteLLMModel.__call__
           |    |
           |    \-- Instruction Non Compliance (LOW):
           |         "Omitting the mandatory <end_plan> tag"
           |
           +-- [Chain] Step 1
           |    |
           |    +-- [LLM] LiteLLMModel.__call__
           |    |
           |    \-- [Tool] FinalAnswerTool
           |
           \-- [LLM] LiteLLMModel.__call__
\end{lstlisting}

\noindent\begin{minipage}{\linewidth}
\captionof{figure}{\small Trace structure for session \texttt{b7f8fcd484777f9d330f24a2ff30dd25}. One planning span is flagged with ``instruction non compliance'' for omitting a formatting tag, despite a perfect \texttt{Overall (GT) = 5.0} score.}
\end{minipage}

\vspace{1em}

\noindent Our Deepchecks \textbf{Instruction Following} metric (score: 3.0) independently identified the same issue: \emph{``...follows all structural and content rules except for omitting the mandatory \texttt{<end\_plan>} tag.''}

Both the human annotation and Deepchecks flagged the same minor formatting issue: a missing tag. The reasoning itself is sound, and the trace achieves a perfect overall score. Yet under existential failure propagation, this trace would be marked as ``having errors'' at the trace level, equivalent to traces with critical reasoning failures.

\vspace{1em}

\noindent\textbf{The problem.} We identified 44 traces in our dataset where \texttt{Overall (GT) $\geq$ 3.5} despite span-level error annotations. Three traces achieve a perfect 5.0 overall score yet still have span-level errors. Recording such minor issues at the span level is valuable for identifying recurring patterns (e.g., which prompts consistently omit formatting tags). However, these annotations should not penalize the trace when they do not materially impact reasoning quality.

\vspace{1em}

\noindent\textbf{Implications.} An important direction for future work is developing smarter aggregation that incorporates severity weighting, outcome-relevance filtering, and reasoning-aware policies leveraging the evaluator's rationales.

\end{document}